\documentclass{article}

\usepackage{arxiv}
\usepackage[utf8]{inputenc} % allow utf-8 input
\usepackage[T1]{fontenc}    % use 8-bit T1 fonts
\usepackage{hyperref}       % hyperlinks
\usepackage{url}            % simple URL typesetting
\usepackage{booktabs}       % professional-quality tables
\usepackage{amsfonts}       % blackboard math symbols
\usepackage{nicefrac}       % compact symbols for 1/2, etc.
\usepackage{microtype}      % microtypography
\usepackage{lipsum}		% Can be removed after putting your text content
\usepackage{graphicx}
\usepackage{natbib}
\usepackage{doi}
\usepackage{wrapfig}
\usepackage{algorithm}
\usepackage[noend]{algpseudocode}
\usepackage[bottom]{footmisc}
\usepackage{amsmath}
\usepackage{xcolor}         % colors

\title{On the Solvability of the {XOR} Problem by Spiking Neural Networks}

\author{ 
{
\hspace{1mm}Bernhard A.~Moser}\thanks{double affiliation: Software Competence Center Hagenberg (SCCH), 4232 Hagenberg, Austria} \\
	Institute of Signal Processing \\
	Johannes Kepler University of Linz\\
	\texttt{bernhard.moser@\{scch.at,jku.at\}} 
	\And
	{\hspace{1mm}Michael Lunglmayr} \\
	Institute of Signal Processing\\
	Johannes Kepler University of Linz, Austria\\
	\texttt{michael.lunglmayr@jku.at} 
	}

\renewcommand{\shorttitle}{\textit{arXiv} On the Solvability of the {XOR} Problem by SNNs}

\hypersetup{
pdftitle={A template for the arxiv style},
pdfsubject={q-bio.NC, q-bio.QM},
pdfauthor={Bernhard A.~Moser, Michael Lunglmayr},
pdfkeywords={Leaky Integrate-and-Fire, Neuromorphic Computing, Spiking Neuron Model, Threshold-based Sampling},
}

\begin{document}
\maketitle

\begin{abstract}
The linearly inseparable XOR problem and the related problem of representing binary logical gates is revisited from the point of view of temporal encoding and its solvability 
by spiking neural networks with minimal configurations of leaky integrate-and-fire (LIF) neurons.
We use this problem as an example to study the effect of different hyper parameters such as information encoding, the number of hidden units 
in a fully connected reservoir, the choice of the leaky parameter and the reset mechanism in terms of 
reset-to-zero and reset-by-subtraction based on different refractory times.
The distributions of the weight matrices give insight into the difficulty, respectively the probability, to find a solution. 
This leads to the observation that zero refractory time together with graded spikes and an adapted reset mechanism, reset-to-mod, makes it possible to realize sparse solutions of a minimal configuration with only two neurons in the hidden layer to resolve all binary logic gate constellations with XOR as a special case.
%However, these solutions are less probable and therefore more difficult to find compared to an increased number of LIF neurons with non-zero refractory time.

\keywords{Spiking Neural Networks (SNNs) \and Leaky Integrate-and-Fire \and Temporal Encoding \and Reservoir Computing }
\end{abstract}

\section{Introduction}
We consider the set $S = \{(0,0), (0,1), (1,0), (1,1)\}$ and all its binary partitions $P = [A, B]$, 
where $P = A \cup B$, $A \cap B = \emptyset$. 
XOR represents the special partition $P_{\mbox{\tiny XOR}} := [\{(0,0), (1,1)\}, \{(0,1), (1,0)\}]$.
Solving the XOR problem refers to specifying a classification model that perfectly separates the subsets of the XOR partition 
$P_{\mbox{\tiny XOR}}$. 
Due to Radon's theorem~\cite{Radon1921MengenKK} for any $d+2$ points in 
$\mathbb{R}^d$ there is a partition into two subsets with intersecting convex hulls.
As a consequence, since for a linear classifier $L$ with threshold $\vartheta$ the related pre-images 
$A :=\{x \in \mathbb{R}^d: L(x)\geq \vartheta\}$ and 
$B :=\{x \in \mathbb{R}^d: L(x)< \vartheta\}$ form a partition of convex sets, $4$ points in $\mathbb{R}^2$ cannot 
be shattered by a linear classifier.
% Thus, the VC dimension of this particular classifier is 3.

\begin{figure}
	\centering
		\includegraphics[width=0.5\textwidth]{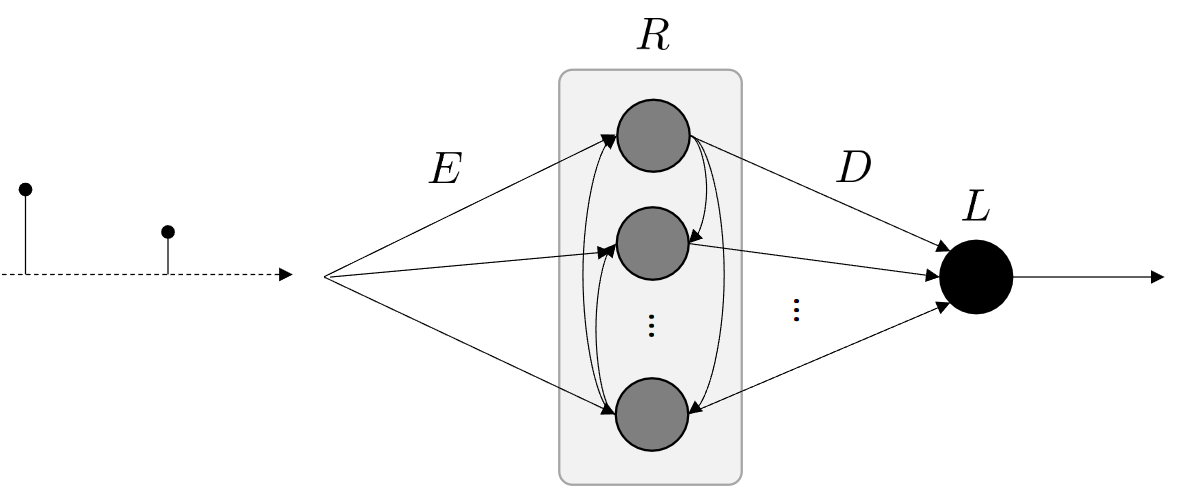}		
  		\caption{Architecture of SNN considered in the paper, consisting of a reservoir $R$ of fully connected hidden 
			leaky integrate-and-fire (LIF) neurons of the same type (same threshold and leaky parameter) with weight matrix $W$ and a linear output classifier $L$. The encoder weights $E = (1, 0, \ldots, 0)$ are fixed, $W$ is generated randomly and the decoder weights $D$ are learned. 
			}
	\label{fig:SNN}
\end{figure}

In this paper we study the problem under which conditions the set $S = \{(0,0), (0,1), (1,0), (1,1)\} \in \mathbb{R}^2$ can be shattered
by a spiking neural network $\mbox{SNN}_{W}$ with a single hidden layer and weight matrix $W$.
Fig.~\ref{fig:SNN} illustrates the architecture of such SNNs.  
We use the XOR problem as a vehicle to get insight into the effect of encoding, the choice of the reset mechanism and the difficulty to find a
solution by considering the distribution of weights $W$. That is, we are looking for an as-simple-as-possible spiking neural network that allows to tune its weight matrix to realize any binary partition of interest.

The paper is structured as follows. Section~\ref{s:RelatedWork} outlines related work on the XOR problem in the context of SNNs.
Section~\ref{s:LIF} recalls the LIF model and the recently introduced reset-to-mod modification.
 Section~\ref{s:XOR} describes the setup of our experiments and discusses its results. 

\section{Related Work}
\label{s:RelatedWork}
Representing a simple non-linear problem  that requires hidden units to transform the input into the desired output, 
the XOR problem is often considered a benchmark problem for testing neural network capabilities in solving more complex problems.
This problem played an remarkable role in the early phase of AI. In their book~\cite{Min69}, the neural network pioneers 
Marvin Minsky and Seymour Papert demonstrated that it is impossible for single-layer perceptrons (also referred to a first generation neural networks) to resolve the XOR problem.
Unfortunately, incorrect citations in connection with these findings contributed to a significant decline in interest of neural network research in the 1970s, the so-called {\it AI winter}. It took another ten years before research in the field of neural networks began to take off in the 1980s~\cite{Sejnowski2018}. In the meanwhile we encounter the so-called third generation of artificial neural networks
in terms of spiking neural networks (SNNs) which are  closer to the biological reference model
by giving time a crucial role in information encoding and dynamics of the network~\cite{Maass1997,bookGerstner2014}.

The neurons in a spiking neural network (SNN) generate action potentials, or spikes, when the internal neuron state variable, called {\it membrane potential}, crosses a threshold. In contrast to conventional neural networks of the first and second generation, this way 
SNNs interconnect neurons that asynchronously  process and transmit spatial-temporal information based on the occurrence of spikes that come from spatially distributed sensory input neurons~\cite{Daya_2001_book,bookGerstner2014}. 

%Due to their particular nature of asynchronous and sparse information processing, SNNs are studied mainly for two reasons: first, as a simplified mathematical model in the context of computational neuroscience aiming at a better understanding of biological neural circuits, and second, as a further  step  towards more powerful but energy-efficient embedded AI edge solutions to process time-varying signals with a wide range of applications including visual processing~\cite{Amir2017,Yousefzadeh2022}, audio recognition~\cite{SNNSound2018}, speech recognition~\cite{SpeechRec2020}, biomedical signal processing~\cite{Hassan2018RealTimeCA} and robotic control~\cite{Kabilan2021,Yamazaki2022}. New application scenarios are emerging in the context of edge AI and federated learning across a physically distributed network of resource-constrained edge devices to collaboratively train a global model while preserving privacy~\cite{YangFedSNN2022}. Of particular interest are applications in the emerging field of brain-computer interfaces which  opens up new perspectives for the treatment of neurological diseases such as Parkinson's disease~\cite{Dethier_2013, Gege2021}.

Inspired from biology different information encoding principles with different characteristics
have been proposed. Two main coding approaches can be distinguished for SNN-based systems: rate coding and temporal coding. For an overview, see, e.g.~\cite{bookGerstner2014,Auge2021}. Rate coding aims to represent the intensity of a variable, e.g. voltage, by means of a spike frequency rate. This principle has been known in neurophysiology for many decades~\cite{AdrianZotterman1926}, so 
it has been experimentally discovered in most sensory systems such as the visual cortex and the motor cortex.
%Due to its robustness and simple mechanism, rate coding has become one of the dominant paradigm in computational neuroscience and artificial neural networks. Its popularity may also be due to its analogy to the activation function of ordinary, non-spike-based artificial neurons. Rate coding is also encountered  in signal processing-based approaches, e.g. delta-sigma modulation based on oversampling to encode a signal into a digital signal with low bit depth and very high sampling frequency~\cite{Danial2019}.  
However, rate coding comes also with drawbacks such as limitations  due to slow information transfer and a long processing time. 
In contrast, temporal coding techniques use the precise timing of and between spikes to encode information. This includes the absolute timing with some reference, the relative timing of spikes triggered by different neurons, or simply the order in which neurons generate certain spikes. 

The various information encoding variants are also taken up to tackle the XOR problem by spiking neural networks~\cite{Bohte2002, Wade2007, ReljanDelaney2017,Enriquez-Gaytan2018, Matsumoto2018, Cyr2020}. So, Bohte et al~\cite{Bohte2002} demonstrate a proof of concept of their SpikeProp algorithm by utilizing temporal encoding. While $0$ is encoded with a {\it late} firing time and $1$ is represented by {\it early} firing time. The related SNN topology consists of three input neurons ($2$ coding neurons and $1$ reference neuron), $5$ hidden neurons and a single output neuron. Due to convergence reasons this model does not allow  a mix of both positive and negative weight. 
Therefore one of the hidden neurons is designed as an inhibitory neuron generating only negative sign spikes. 
In contrast, other authors such as~\cite{Wade2007,ReljanDelaney2017,Enriquez-Gaytan2018,Cyr2020} 
utilize rate encoding by representing $0$ by spike trains of some frequency, e.g. 50Hz, and $1$ by another frequency, e.g., 100Hz.
By mimicking logic gates, in \cite{Wade2007} the SNN topology for the XOR problem consists of two inputs, $2$ hidden layers with $4$ neurons each, and $2$ output neurons, where the the first hidden layer is partially connected, based on neurons that are designed to respond on selected frequency ranges, resulting in two active neurons for any $0$-$1$ combination. 
In the same spirit, also~\cite{ReljanDelaney2017} mimics the functionality of logic gates but by 
utilizing additionally receptive fields between the LIF neurons to realize selective responses input frequencies.
The resulting feed-forward SNN also consists of $2$ input neurons and $4$ LIF neurons in a hidden layer
together with additional $2\times8$ receptive fields (RF) to filter the states $(0,0)$, $(0,1)$, $(1,0)$ and $(1,1)$, 
and two output neurons for $0$ and $1$, where the final decision is based on the winner-takes-all principle.
\cite{Cyr2020} proposes four main layers of LIF neurons based on spike timing-dependent plasticity (STDP) as learning rule.
%The proposed models also differ in terms of the used transfer function. 
While leaky integrate-and-fire (LIF) is the simplest neuron model for SNNs, 
also more advanced neuron models such as the Izhikevich neuron are used.
Again using rate encoding, \cite{Enriquez-Gaytan2018} studies the XOR problem by means of 
a feed-forward $2-2-1$ SNN architecture based on Izhikevich neurons. Its related $16$ weights are 
found by genetic and evolutionary algorithms.

\section{LIF Model and Preliminaries}
\label{s:LIF}
The leaky-integrate-and-fire neuron model (LIF) with leaky  parameter $\alpha>0$ and threshold $\vartheta>0$ uses 
integration which determines a recursive procedure to turn a signal $f$ into a spike train $\eta(t) = \sum_k s_k\delta(t - t_k)$, where $s_k \in \mathbb{R}$ denotes the amplitude of the spike at time $t_k$. 
The time points $t_{k}$ are recursively given by
\begin{equation}
\label{eq:LIFsample}
t_{k+1} := \inf\left\{T\geq t_k + t_r: \mathcal{T}\left[\int_{t_k}^{T} e^{-\alpha (t_{k+1} -  t)} \big(f(t) + r_k \delta(t-t_k)\big) dt\right] \geq \vartheta\right\},
\end{equation}
where $\mathcal{T}[x] = x$ is either the identity (only positive threshold), or the modulus, $\mathcal{T}[x] = |x|$ (positive and threshold), 
$t_r\geq 0$ is the refractory time and $T=t_{k+1}$ is the first time point after $t_{k}$ that causes the integral in 
(\ref{eq:LIFsample}) to violate the sub-threshold condition $|\int_{t_k}^{T} e^{-\alpha (t_{k+1} -  t)} f(t) dt | < \vartheta$.
The term $r_k \delta(t-t_k)$ refers to the reset of the membrane potential in the moment a spike has been triggered.
In the standard definition of LIF for discrete spike trains, see~\cite{bookGerstner2014}, the reset is defined as the membrane potential that results from subtracting the threshold if the membrane's potential reaches the positive threshold level $+\vartheta$, or adding $\vartheta$ to the membrane's potential if a spike is triggered at the negative threshold level $-\vartheta$. 
In the case of bounded $f$ the integral $g(t):= \int_{t_k}^t e^{-\alpha (t_{k+1} -  t)} f(t) dt$ is changing continuously in $t$ 
so that the threshold level in (\ref{eq:LIFsample}) is exactly hit. 
Consequently the resulting reset amounts to zero, i.e., $r_k = 0$ and the resulting amplitude $s_k$ of the triggered spike is defined accordingly, i.e., $s_k = +\vartheta$, when the positive threshold value is reached, and $s_k = -\vartheta$ when the negative threshold value is reached.
For a mathematical analysis and a discussion of how to define the reset $r_k$  in the presence of Dirac impulses see~\cite{moserSNNAlexTop}.

Injecting weighted Dirac pulses the neuron's potential will show discontinuous jumps, and different reset variants are reasonable from an algorithmic point of view.
Beyond the prevalent variants of {\it reset-to-zero} and {\it reset-by-subtraction}, see e.g.~\cite{snnTorch2021}, 
recently we introduced {\it reset-to-mod} as a third option, see~\cite{moserSNNAlexTop}. 
{\it reset-to-zero} means that the neuron's potential is reinitialized to zero after firing, while {\it reset-by-subtraction} subtracts the $\vartheta$-potential $u_{\vartheta}$ from the membrane's potential that triggers the firing event. The third variant,  {\it reset-to-mod}, can be understood as instantaneously cascaded application of {\it reset-by-subtraction} according to the factor $n$ by which the membrane's potential $u$ exceeds the threshold, i.e.
$u = n \vartheta + r$, $r \in ]-\vartheta, \vartheta[$. This means that {\it reset-to-mod} is the limit case of {\it reset-by-subtraction} with the refractory time $t_r$ approaching to zero. In this case the residuum $r$ results from a modulo computation and the amplitude of the triggered spike is set to 
$n\, \vartheta$. 

As listed in Table~\ref{tab:0}, in total we get $6$ LIF neuron model variants depending on the choice of thresholding (only positive, or positive and negative) and the reset variants {\it reset-to-mod}, {\it reset-by-subtraction}  or {\it reset-to-zero}.
\begin {table}[h]
\caption {LIF Spiking Neuron Model Variants} 
\label{tab:0} 
\begin{center}
\begin{tabular}{lcc}
\hline
  Model  & Thresholding &   Reset  \\
\hline
  Symmetric Reset-to-Mod (SRM)  & positive and negative &  reset-to-mod \\
  Symmetric Reset-by-Sub (SRS) &  positive and negative  & reset-by-subtraction\\
  Symmetric Reset-to-Zero (SRZ) &  positive and negative &  reset-to-zero   \\
  Positive Reset-to-Mod (PRM) & only positive &  reset-to-mod  \\
  Positive Reset-by-Sub (PRS) &  only positive &  reset-by-subtraction  \\
  Positive Reset-to-Zero (PRZ) &  only positive &  reset-to-zero   \\
\hline
\end{tabular}
\end{center}
\end{table}

%
%With discretization time interval $\Delta$ and discrete time points at $t_k := k\,\Delta$, 
%the signal $f = f(t)$ becomes a sequence $f_k := f(t_k)\, 1_{(k-1, k]}$. 
%Then {\it reset-to-mod} turns Eq.~(\ref{eq:LIFsample}) into a recursion that maps $f$ to the spike train 
%$s = \sum_k s_k \, \delta(t - t_k)$ given by
%\begin{eqnarray}
%\label{eq:DefLIFd}
%z_1 & := & f_1, 																	\nonumber \\
%s_1 & := & q_{\vartheta}(z_1), 															\nonumber \\
%z_{k+1} & := & f_{k+1} + \beta\, (z_k - s_k),		\nonumber \\
%s_{k+1} & := & q_{\vartheta}(z_{k+1}),
%\end{eqnarray}
%where $\beta := e^{-\alpha} \in [0,1]$ and $q_{\vartheta}$ denotes the standard quantization by truncation with respect to the regular %grid 
%$\vartheta\, \mathbb{Z}$, i.e., a point $x$ is mapped to the next grid point closest to zero.

\section{Resolving Binary Logical Gates by SNNs}
\label{s:XOR}
In contrast to the related work outlined in Section~\ref{s:RelatedWork}, our model consists only of a single input neuron and a single output neuron as illustrated in Fig.~\ref{fig:SNN}. 
The hidden layer is realized by a reservoir of $N$ LIF neurons with randomly generated weights~\cite{Rahimi2008}. 
The decision is realized by a classical perceptron by summing up the weighted output spike trains 
$\psi_k$, $k = 1,\ldots, N$ at neuron $L$ in Fig.~\ref{fig:SNN}.
%i.e., $\sum_k d_k \psi_k(j)$
%= \sum_j s^k_j \delta (t-t_j^k)$
Note that the existence whether there is a solution or not can be checked by a linear program. 
If $v^A_i$ denotes the vector of resulting sums for each output channel for the $i$-th input from the $A$ class, and 
accordingly, $v^B_j$ for $j$-th input from class $B$, then there the classes $A$ and $B$ can be separated linearly if and only if
there is a vector $D$ of decoder weights and a threshold $\vartheta$
such that for all $i, j$ we have $\langle v_i^A, D\rangle \geq \vartheta$ and $\langle v_j^B, D\rangle < \vartheta$, where 
$\langle x, y\rangle = \sum_i x_i\, y_i$ denotes the standard inner product.
This solvability problem can equivalently be decided by a homogenous problem with threshold $0$ by checking the existence of $\widetilde{D}$ such that
$\langle \widetilde{v}_i^A, \widetilde{D}\rangle \geq 0$ and $\langle \widetilde{v}_j^B, \widetilde{D}\rangle < 0$ for $i$ and $j$,
where $\widetilde{v}_i^A$ and $\widetilde{v}_j^B$ result from $v_i^A$, resp. $v_j^B$, by adding the additional coordinate $-1$.
In turn, this problem can be equivalently decided by checking whether the set constituted of all $\widetilde{v}_i^A$ and $-\widetilde{v}_j^B$ can be linearly separated from the origin $0$, i.e., whether the convex hull of the points $\widetilde{v}_i^A$ and $-\widetilde{v}_j^B$ contains $0$ or not. 
This can be done by a linear program to solve for $x = (x_1, \ldots,x_n)$ satisfying 
$P x= 0$, $\sum_k x_k = 1$ and $x_k\geq 0$, where $P$ is the matrix containing the points as column vectors, see, e.g.,~\cite{Matouek2006}.

\begin{figure}
	\centering
		\includegraphics[width=0.8\textwidth]{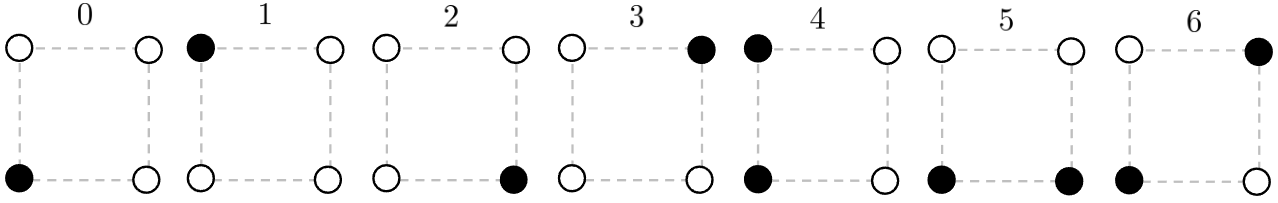}		
  		\caption{Enumeration of all binary logical gates where XOR is represented by case $6$.}
	\label{fig:gates}
\end{figure}

\begin{figure}
	\centering
		\includegraphics[width=0.45\textwidth]{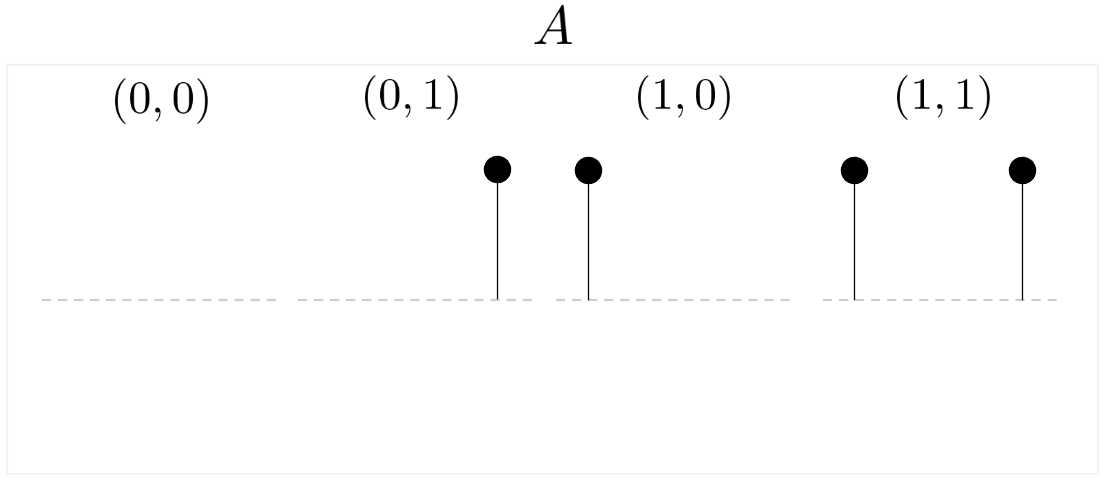}		
		\includegraphics[width=0.45\textwidth]{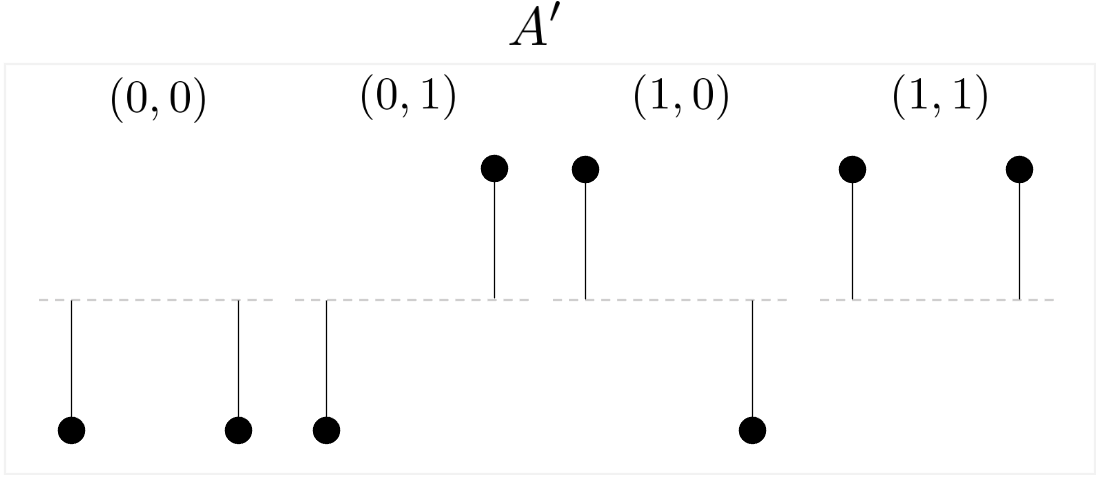}		
		\includegraphics[width=0.45\textwidth]{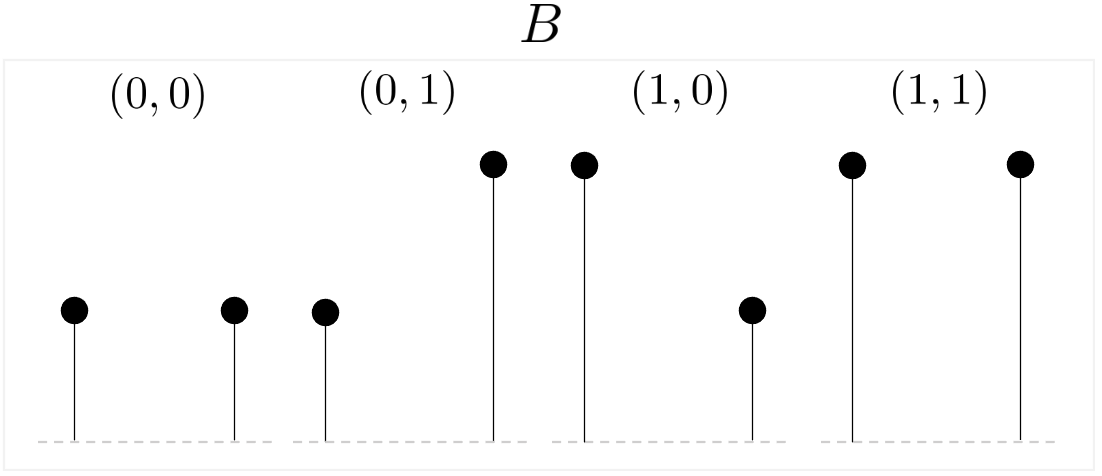}		
		\includegraphics[width=0.45\textwidth]{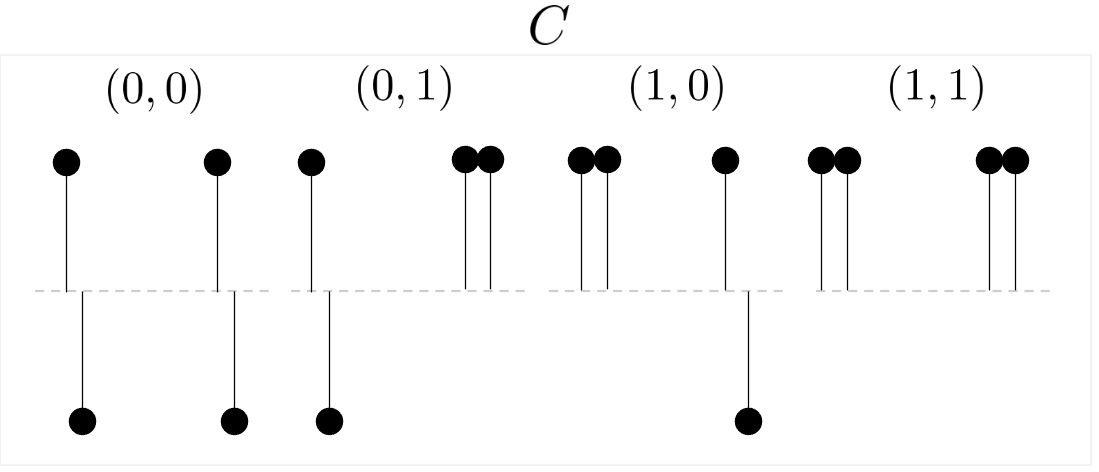}		
  		\caption{Variants $A$, $A'$, $B$ and $C$ for temporal encoding of logical gates by means of mixed positive and negative graded spikes.}
	\label{fig:Encodings}
\end{figure}

Our main objective is to investigate the effect of various hyper parameters on the distribution of solutions in the space of weights, 
that is how difficult it is to find a solution. Besides the XOR problem, we also consider all the other constellations of binary logical gates as illustrated in Fig.~\ref{fig:gates}.
For this we consider temporal encoding in different constellations, also allowing mixed positive, negative spikes and spikes with different amplitudes (grades), see Fig.~\ref{fig:Encodings}. The weights for the fully connected reservoir of LIF neurons are generated randomly based on uniform sampling 
the interval $[-1,1]$ with discretization of $0.1$. The probability evaluations are based on $100$ runs in each considered constellation.
Note that the variants $A$, $A'$ and $B$ of Fig.~\ref{fig:Encodings} represent $4$ points in $2$-dimensional space, whereas this is not the case for $C$.
Therefore, only $A$ or $B$ are representation to which Radon's theorem of non-separability applies. 
Variant $C$ circumvents the problem by increasing the dimensionality of the space from $2$ to $4$. 
Dimensionality enlargement by additional spikes might ease the problem as demonstrated in Table~\ref{tab:3}, but at the cost of sparseness.
For $\beta=1$ for all LIF variants SRM, SRS, SRZ, PRM, PRS, PRZ one can find solutions, where  the probability to find a solution 
is greater the $10\%$ for SRM and SRS.  For $\beta=0.5$ only for SRM and SRS there are solutions.  

While encoding $A$ does not work, the results for encoding variant $B$ and leaky parameters $\beta=1$, resp. $\beta=0.5$, are shown in Table~\ref{tab:1}.
Interestingly, our recently introduced {\it reset-to-mod} variant in terms of SRM (symmetric reset-to-mod) and PRM (positive reset-to-mod)
gives the highest probability to find a solution, particularly for PRM with $\beta=0.5$. Table~\ref{tab:11} and Table~\ref{tab:12}
show the related mean and standard deviation of the $l_1$-norm of the output spike train.  
For PRM with $\beta=0.5$ we obtain the sparsest solutions for all gates.

\begin {table}[h]
\caption {Solvability probability for encoding B, $E=(1,0)$, $\beta=1$ (left), $\beta=0.5$ (right)} 
\label{tab:1} 
\begin{center}
\begin{tabular}{crrrrrr}
\hline
  Gate  &   SRM &   SRS &   SRZ &    PRM &   PRS &   PRZ \\
\hline
  0 & 30.0 & 27.5 &  0.0 &  49.5 &  49.5 &   0.0 \\
  1 & 2.5 &  2.5 &  0.0 &   1.0 &   1.0 &   0.0 \\
  2 & 3.0 &  2.5 &  0.0 &   1.0 &   1.5 &   0.0 \\
  3 & 29.5 & 28.5 &  0.0 &  50.0 &  49.5 &   0.0 \\
  4 & 1.5 &  3.0 &  0.0 &   1.0 &   5.0 &   0.0 \\
  5 & 4.0 &  4.5 &  0.0 &   0.5 &   1.5 &   0.0 \\
  6 & 2.0 &  1.0 &  0.0 &   1.0 &   0.5 &   0.0 \\
\hline
\end{tabular}
\begin{tabular}{crrrrrr}
\hline
  Gate  &   SRM &   SRS &   SRZ &   {\bf PRM} &   PRS &   PRZ \\
\hline
  0 & 63.5 & 22.5 &  0.0 & {\bf 93.0} &  25.0 &   0.0 \\
  1 & 20.0 &  0.0 &  0.0 & {\bf  11.5} &   0.0 &   0.0 \\
  2 & 14.5 &  0.0 &  0.0 & {\bf  11.5} &   0.0 &   0.0 \\
  3 &  70.0 & 22.5 &  0.0 & {\bf  95.5} &  25.0 &   0.0 \\
  4 &  10.5 &  2.0 &  0.0 & {\bf   2.5} &   1.0 &   0.0 \\
  5 &   5.0 &  2.0 &  0.0 & {\bf   2.5} &   1.0 &   0.0 \\
  6 &  7.5 &  0.0 &  0.0 &  {\bf  3.0} &   0.0 &   0.0 \\
\hline
\end{tabular}
\end{center}
\end{table}

\begin {table}[h]
\caption {Mean of $l_1$-norm of output spike train of Table~\ref{tab:1}, $\beta=1$ (left), $\beta=0.5$ (right); "-" means not computable due to no spikes available.} 
\label{tab:11} 
\begin{center}
\begin{tabular}{crrrrrr}
\hline
  Gate  &   SRM &   SRS &   SRZ &     PRM &   PRS &   PRZ \\
\hline
 0 & 12.7 & 12.8 &  - &    4.4 &   6.6 &   - \\
  1 & 32.2 & 37.0 &  - &   5.0 &  18.0 &   - \\
  2 & 41.0 & 15.6 &  - &   5.0 &  15.0 &   - \\
  3 &  5.6 & 10.2 &  - &  3.4 &  4.5 &   - \\
  4 & 31.7 & 50.3 &  - &  6.0 &  16.2 &   - \\
  5 & 17.0 & 27.3 &  - &  6.0 &  16.0 &   - \\
  6 & 18.8 & 36.5 &  - &  6.0 &  8.0 &   - \\
\hline
\end{tabular}
\begin{tabular}{crrrrrr}
\hline
  Gate  &    SRM &   SRS &   SRZ &    {\bf PRM} &   PRS &   PRZ \\
\hline
  0 & 5.9 &   5.2 &  - &   {\bf 3.2} &   4.9 &   - \\
  1 & 24.0 & -   &  - &   {\bf 3.2} & -   &   - \\
  2 & 11.4 & -   &  - &   {\bf 3.6} & -   &   - \\
  3 & 3.1 &   3.5 &  - &   {\bf 2.2} &   3.4 &   - \\
  4 & 42.0 &   6.5 &  - &  {\bf  6.0} &   7.0 &   - \\
  5 & 7.3 &   6.5 &  - &   {\bf 8.8} &   7.0 &   - \\
  6 & 3.8 & -   &  - &   {\bf 3.8} & -   &   - \\
\hline
\end{tabular}
\end{center}
\end{table}

\begin {table}[h]
\caption {Standard deviation of $l_1$-norm w.r.t Table~\ref{tab:11}} 
\label{tab:12} 
\begin{center}
\begin{tabular}{crrrrrr}
\hline
  Gate  &   SRM &   SRS &   SRZ &    PRM &   PRS &   PRZ \\
\hline
 0 & 24.4 & 21.3 &  - &    4.3  &   5.5 &   - \\
  1 & 47.9 & 39.6 &  - &    1.0  &   1.0 &   - \\
  2 & 47.4 &  4.1 &  - &   1.0  &   4.3 &   - \\
  3 &  9.2 & 25.0 &  - &   3.5  &   3.7 &   - \\
  4 & 36.3 & 45.6 &  - &   0.0  &   3.4 &   - \\
  5 & 25.1 & 31.8 &  - &   0.0  &   7.0 &   - \\
  6 & 22.7 & 28.5 &  - &   1.0  &   0.0 &   - \\
\hline
\end{tabular}
\begin{tabular}{crrrrrr}
\hline
  Gate  &   SRM &   SRS &   SRZ &    {\bf PRM} &   PRS &   PRZ \\
\hline
  0 &  16.6 &   1.8 &  - &   {\bf  2.1} &   1.6 &   - \\
  1 & 57.7 & -   &  - &   {\bf 0.4} & -   &   - \\
  2 & 33.9 & -   &  - &   {\bf  1.7} & -   &   - \\
  3 &  1.5 &   0.9 &  - &   {\bf  1.3} &   0.8 &   - \\
  4 & 75.1 &   0.9 &  - &   {\bf 2.8} &   1.0 &   - \\
  5 &  3.2 &   0.9 &  - &   {\bf 3.7} &   1.0 &   - \\
  6 &  1.0 & -   &  - &   {\bf 1.5} & -   &   - \\
\hline
\end{tabular}
\end{center}
\end{table}

\begin {table}[h]
\caption {Solvability probability for encoding C, $E=(1,0)$, $\beta=1$ (left), $\beta=0.5$ (right)} 
\label{tab:3} 
\begin{center}
\begin{tabular}{crrrrrr}
\hline
  Gate  &   SRM &   SRS &   SRZ &    PRM &   PRS &   PRZ \\
\hline
  0 & 61.0 & 72.0 & 66.0 &  41.0 &  41.0 &  40.0 \\
  1 & 19.0 & 30.0 & 27.0 &   3.0 &   3.0 &   6.0 \\
  2 & 24.0 & 29.0 & 22.0 &   7.0 &   7.0 &   7.0 \\
  3 & 57.0 & 66.0 & 70.0 &  59.0 &  59.0 &  40.0 \\
  4 & 34.0 & 41.0 & 27.0 &  62.0 &  61.0 &  38.0 \\
  5 & 14.0 & 17.0 &  8.0 &   5.0 &   5.0 &   3.0 \\
  6 & 13.0 & 17.0 &  8.0 &   6.0 &   6.0 &   3.0 \\
\hline
\end{tabular}
\begin{tabular}{crrrrrr}
\hline
  Gate  &  {\bf SRM} &   {\bf SRS} &   SRZ &   PRM &   PRS &   PRZ \\
\hline
  0 & {\bf 87.0} & {\bf 90.0} & 88.0 &  39.0 &  39.0 &  39.0 \\
  1 & {\bf 15.0} & {\bf 15.0} & 14.0 &   5.0 &   5.0 &   5.0 \\
  2 & {\bf 6.0} & {\bf 6.0} &  6.0 &   0.0 &   0.0 &   0.0 \\
  3 & {\bf 95.0} & {\bf 98.0} & 95.0 &  44.0 &  44.0 &  44.0 \\
  4 & {\bf 39.0} & {\bf 39.0} & 37.0 &  44.0 &  44.0 &  44.0 \\
  5 & {\bf 2.0} &  {\bf 2.0} &  0.0 &   0.0 &   0.0 &   0.0 \\
  6 & {\bf 11.0} & {\bf 11.0} &  8.0 &   5.0 &   5.0 &   5.0 \\
\hline
\end{tabular}
\end{center}
\end{table}

\begin {table}[h]
\caption {Mean of $l_1$-norm of output spike train of Table~\ref{tab:3}, $\beta=1$ (left), $\beta=0.5$ (right)} 
\label{tab:31} 
\begin{center}
\begin{tabular}{crrrrrr}
\hline
  Gate  &   SRM &   SRS &   SRZ &     PRM &   PRS &   PRZ \\
\hline
 0 &  8.2 & 11.6 &  6.7 &   7.5 &   8.3 &   4.5 \\
  1 &  6.2 & 14.7 &  5.1 &  12.3 &  13.3 &   5.8 \\
  2 & 13.8 &  6.5 &  5.0 &   9.4 &   9.9 &   5.7 \\
  3 &  4.3 &  4.1 &  5.2 &   3.5 &   3.5 &   3.2 \\
  4 &  7.5 & 14.1 & 12.1 &   5.7 &   5.9 &   4.4 \\
  5 & 21.9 & 25.4 & 98.6 &   9.8 &  10.4 & 190.0 \\
  6 &  4.5 &  4.1 & 26.4 &   4.7 &   4.7 &   3.3 \\
\hline
\end{tabular}
\begin{tabular}{crrrrrr}
\hline
  Gate  &    {\bf SRM} &  {\bf  SRS} &   SRZ &    PRM &   PRS &   PRZ \\
\hline
  0 &  {\bf 3.7} &  {\bf 3.6} &   4.0 &   5.4 &   5.4 &   4.0 \\
  1 & {\bf  7.7} &  {\bf 7.7} &   5.7 &   6.8 &   6.8 &   6.0 \\
  2 & {\bf  8.0} &  {\bf 8.0} &   5.0 & -   & -    & -    \\
  3 & {\bf  3.5} &  {\bf 3.5} &   3.2 &   3.8 &   3.8 &   3.1 \\
  4 &  {\bf 6.2} &  {\bf 6.2} &   4.5 &   5.5 &   5.5 &   4.2 \\
  5 & {\bf 12.0} & {\bf 12.0} & -    & -    & -    & -    \\
  6 & {\bf  5.1} & {\bf  5.1} &   4.1 &   4.4 &   4.4 &   4.0 \\
\hline
\end{tabular}
\end{center}
\end{table}

\begin {table}[h]
\caption {Standard deviation of $l_1$-norm w.r.t Table~\ref{tab:31}} 
\label{tab:32} 
\begin{center}
\begin{tabular}{crrrrrr}
\hline
  Gate  &     SRM &    SRS &   SRZ &    PRM &   PRS &   PRZ \\
\hline
 0 & 13.4 &    24.8 & 21.7 &   4.2 &   5.2 &   1.1 \\
  1 &   3.1 & 33.0 &  1.6 &   3.3 &   3.1 &   1.1 \\
  2 &  33.6 &  3.6 &  1.6 &   3.8 &   4.1 &   1.0 \\
  3 &   2.4 &  2.3 & 11.5 &   1.7 &   1.7 &   0.5 \\
  4 &   4.0 & 28.3 & 33.2 &   3.8 &   3.6 &   1.1 \\
  5 &  42.3 & 41.1 & 91.4 &   3.2 &   3.1 &   0.0 \\
  6 &  1.5 &  1.6 & 58.8 &   1.5 &   1.5 &   0.5 \\
\hline
\end{tabular}
\begin{tabular}{crrrrrr}
\hline
  Gate  &  {\bf  SRM} &   {\bf SRS} &   SRZ &     PRM &   PRS &   PRZ \\
\hline
  0 &  {\bf 2.5} & {\bf  2.5} &  10.6 &   1.6 &   1.6 &   0.0 \\
  1 &  {\bf 3.0} & {\bf  3.0} &   1.3 &   1.6 &   1.6 &   0.0 \\
  2 &  {\bf 3.5} & {\bf  3.5} &   1.5 & -    & -    & -    \\
  3 &  {\bf 1.0} & {\bf  1.0} &   0.5 &   0.8 &   0.8 &   0.3 \\
  4 &  {\bf 2.3} & {\bf  2.3} &   1.0 &   1.6 &   1.6 &   0.6 \\
  5 &  {\bf 2.0} & {\bf  2.0} & -    & -    & -    & -    \\
  6 &  {\bf 1.4} & {\bf  1.4} &   0.3 &   0.8 &   0.8 &   0.0 \\
\hline
\end{tabular}
\end{center}
\end{table}

\section{Conclusion}
\label{s:Conclusion}
In this paper we study the problem to realize the decision problems of binary 
logical gates by means of spiking neural networks based on temporal encoding.
It turns out that the choice of hyper parameters in terms of leaky parameter and the design of
the reset mechanism  in combination with the temporal encoding is crucial. 
In contrast to the standard setting of {\it reset-by-subtraction} we consider 
also {\it reset-to-mod} which can be understood as an instantaneous charge-discharge event with zero net voltage.
It is shown that a temporal encoding of $0$ and $1$ based on graded spikes in combination with
{\it reset-to-mod} and a reservoir of $2$  fully interconnected LIF neurons provides the sparsest solution 
and that the weights can be found by uniform random initialization with a success rate of at least $3\%$, in our experiments $3$ out of $100$ runs.
In future research we will also consider a LIF neuron as output layer which requires a generalization of the outlined solvability criterion 
based on the convex hull argument.

\section*{Acknowledgements}
%This work was supported (1) by the 'University SAL Labs' initiative of Silicon Austria Labs (SAL) and its Austrian partner universities for applied fundamental research for electronic based systems, (2) by Austrian ministries BMK, BMDW, and the State of Upper-Austria in the frame of SCCH, part of the COMET Programme managed by FFG, and (3) by the {\it NeuroSoC} project funded under the Horizon Europe Grant Agreement number 101070634.
This work was supported (1) by the 'University SAL Labs' initiative of Silicon Austria Labs (SAL) and its Austrian partner universities for applied fundamental research for electronic based systems, (2) by Austrian ministries BMK, BMDW, and the State of Upper-Austria in the frame of SCCH and its project S3AI, part of the COMET Programme managed by FFG, and (3) by the {\it NeuroSoC} project funded under the Horizon Europe Grant Agreement number 101070634.

% ****************************************************************************
% BIBLIOGRAPHY AREA
% ****************************************************************************

%\bibliographystyle{plain}
%\bibliographystyle{unsrtnat}
%\bibliographystyle{splncs04}
%unsrtnat
%\bibliography{references}

\end{document}